\def\year{2019}\relax
\documentclass[letterpaper]{article} 
\usepackage{aaai19}  
\usepackage{times}  
\usepackage{helvet}  
\usepackage{courier}  
\usepackage{url}  
\usepackage{graphicx}  
\usepackage{amsmath}
\usepackage{amssymb}
\usepackage{multirow}
\usepackage{epsfig}
\usepackage{algorithm}
\usepackage{algorithmic}
\usepackage{capt-of}
\usepackage{pdfpages}
\usepackage{enumerate}
\usepackage{footnote}
\usepackage{multirow}
\usepackage{makecell}
\usepackage{amsfonts}

\begin{document}
%
\title{PortraitGAN for Simultaneous Emotion and Modality Manipulation}
\author{Jiali Duan\textsuperscript{1} \hspace{.04in} Xiaoyuan Guo\textsuperscript{2} \hspace{.04in} Yuhang Song\textsuperscript{1} \hspace{.04in} Chao Yang\textsuperscript{1} 
\vspace{.05in}
\hspace{.04in} C.C-Jay Kuo\textsuperscript{1} \\
\textsuperscript{1}\thinspace University of Southern California \hspace{.1in} \vspace{.03in}\textsuperscript{2}\thinspace Emory University\\
}
\maketitle
\begin{abstract} 
Modern mobile apps have made possible the transfer of style and discrete facial-attributes, but it's more desirable if we can continuously manipulate our face however we want and convert across modalities at the same time. In this paper, we propose the first model that supports continuous edits and multi-modality portrait manipulation in a single framework using adversarial learning. Specifically, we adapt cycle-consistency into the conditional setting by leveraging additional facial landmarks information. This has two effects: first cycle mapping induces bidirectional manipulation and identity preserving; second pairing samples from different modalities can thus be utilized. To ensure high-quality synthesis, we adopt texture-loss that enforces modality consistency and multi-level adversarial supervision that facilitates gradient flow. Quantitative and qualitative experiments show the effectiveness of our framework in performing flexible and multi-modality portrait manipulation with photo-realistic effects.
\end{abstract}

\noindent Portrait manipulation has exerted an universal appeal, manifested by the abundance of filtered selfies and photos with varing effects such as expression editing (smiling, crying, angry, making eyes bigger etc) or style transfers (silhouette, photo sketch, skin whitening, haze removal etc). The former is facial manipulation at the discrete level while the latter can be generalized as conversion among modalities. Here we define modality $\mathcal{X}_{1}$ and $ \mathcal{X}_{2}$ as pictures having figures of the same emotion but with different styles, textures or artistic expressions (Photo$\leftrightarrow$Stylized, Photo$\leftrightarrow$Cartoonized, Original$\leftrightarrow$Beautified). It would be more preferable if users can incorporate multiple manipulations into a single operation. Therefore, our goal in this paper is to extend discrete facial manipulation into continuous domain and to simultaneously perform modality transformation using a single framework. At least three advantages would emerge from this design: (1) Repetitive operations eliminated once target modality is fixed; (2) Its bidirectional transformation could facilitate face detection algorithm when user's portrait is in an artistic domain; (3) Our design is bidirectional, which allows for inverse conversion $\mathcal{X}_{2}\rightarrow \mathcal{X}_{1}$. In other words, our model is capable of continuing manipulation at any given domain without the need to start from scratch (at a particular domain). 

Cycle-consistency is not a new idea~\cite{zhu2017unpaired,yi2017dualgan} and it was first proposed for unpaired image translation between two domains. In this paper, we extend the idea into a conditional setting by leveraging additional facial landmark information, which is capable of capturing intricate expression changes. Advantages that arise with this simple yet straight-forward modifications include: \textit{First}, cycle mapping can effectively prevent many-to-one mapping~\cite{zhu2017unpaired,zhu2017toward} also known as mode-collapse~\cite{salimans2016improved}. In the context of face/pose manipulation, cycle-consistency also induces identity preserving and  bidirectional manipulation, whereas previous method~\cite{averbuch2017bringing} assumes neutral face to begin with or is unidirectional~\cite{ma2017pose,pumarola2018unsupervised} which manipulate in the same domain. \textit{Second}, face images of different styles or artistic forms~\cite{liao2017visual} are considered different modalities and current landmark detector will not work on those stylized images. With our design, we can pair samples from multiple domains and translate between each pair of them, thus enabling landmark extraction indirectly on portraits of different modalities. Our framework can also be extended to any desired target domains such as makeups/de-makeups, photo/caricature, aging manipulation etc, once corresponding data pairs are collected. 


To synthesize 512x512 images of photo-realistic quality, we propose multi-level adversarial supervision where synthesized images at different resolution are propagated and combined before being fed into multi-level discriminators. Second, to avoid texture inconsistency and artifacts during translation between different domains, we integrate Gram matrix~\cite{gatys2016image} as a measure of texture discrepancy into our model as it is differentiable and can be trained end-to-end using back propagation. We carefully evaluate the role of each component of our approach by conducting ablation study.

Extensive evaluations have shown both quantitatively and qualitatively that our method is comparable or superior to state-of-the-art generative models in performing high-quality portrait manipulation. Our model is bidirectional, which circumvents the need to start from a neutral face or a fixed domain. This feature also ensures stable training, identity preservation and is easily scalable to other desired domain manipulations. In the following section, we review related works to ours and point out the differences.

\section{Related Work}
\paragraph{Face editing} Face editing or manipulation is a widely studied area in the field of computer vision and graphics, including face morphing~\cite{blanz1999morphable}, expression edits~\cite{sucontphunt2008interactive,lau2009face}, age progression~\cite{kemelmacher2014illumination}, facial reenactment~\cite{blanz2003reanimating,thies2016face2face,averbuch2017bringing}. However, these models are designed for a particular task and rely heavily on domain knowledge and certain assumptions. For example, ~\cite{averbuch2017bringing} assumes neutral and frontal faces to begin with while~\cite{thies2016face2face} employs 3D model and assumes the availability of target videos with variation in both poses and expressions. Our model differs from them as it is a data-driven approach that does not require domain knowledge, designed to handle general face manipulations.

\vspace{-3mm}
\paragraph{Image translation} 
Our work can be categorized into image translation with generative adversarial networks~\cite{isola2017image,chen2017photographic,hoffman2017cycada,liu2017unsupervised,yi2017dualgan,wang2017high}, whose goal is to learn a mapping $G:\mathcal{X} \rightarrow  \widehat{\mathcal{Y}}$ that induces an indistinguishable distribution to target domain $\mathcal{Y}$, through adversarial training. For example, Isola et al.~\cite{isola2017image} takes image as a condition for general image-to-image translation trained on paired samples. Later, Zhu et.al~\cite{zhu2017unpaired} builds upon~\cite{isola2017image} by introducing cycle-consistency loss to obviate the need of matched training pairs. In addition, it alleviates many-to-one mapping during training generative adversarial networks also known as mode collapse. Inspired by this, we integrate this loss into our model for identity preservation between different domains.

Another seminal work that inspired our design is StarGAN~\cite{choi2017stargan}, where target facial attributes are encoded into a one-hot vector. In StarGAN, each attribute is treated as a different domain and an auxiliary classifier used to distinguish these attributes is essential for supervising the training process. Different from StarGAN, our goal is to perform continuous edits in the pixel space that cannot be enumerated with discrete labels. This implicitly implies a smooth and continuous latent space where each point in this space encodes meaningful axis of variation in the data. We treat different style modalities as domains in this paper and use two words interchangeably. In this sense, applications like beautification/de-beautification, aging/younger, with beard/without beard can also be included into our general framework. We compare our approach against CycleGAN~\cite{zhu2017unpaired} and StarGAN~\cite{choi2017stargan} during experiments and illustrate in more details about our design in the next section.

\vspace{-3mm}
\paragraph{Pose image generation} There are works that use pose as condition in the task of person re-identification for person image generation~\cite{walker2017pose,lassner2017generative,siarohin2017deformable,pumarola2018unsupervised}. For example~\cite{ma2017pose} concatenates one-hot pose feature maps in a channel-wise fashion to control pose generation similar to~\cite{reed2016learning}, where keypoints and segmentation mask of birds are used to manipulate locations and poses of birds. To synthesize more plausible human poses, Siarohin et.al~\cite{siarohin2017deformable} develop deformable skip connections and compute a set of affine transformations to approximate joint deformations. These works share some similarity with ours as both facial landmark and human skeleton can be seen as a form of pose representation. However, the above works deal with manipulation in the original domain and does not preserve identity. 

\vspace{-3mm}
\paragraph{Style transfer} Neural style transfer was first proposed by Gatys et al.~\cite{gatys2016image}. The idea is to preserve content from the original image and mimic ``style'' from the reference image. We adopt Gram matrix in our model to enforce pattern consistency and replace L-BFGS iteration with back propagation for end-to-end training. Also, considering the lack of groundtruth data of many face manipulation tasks, we apply a fast neural style transfer algorithm~\cite{johnson2016perceptual} to generate pseudo targets for multi-modality manipulations. Note that our model is easily extensible to any desired target domains with current design unchanged. \\

\noindent We also notice two concurrent works similar to ours during preparation of this paper~\cite{pumarola2018ganimation,chan2018everybody}, where GANimation~\cite{pumarola2018ganimation} proposes to address continuous animation conditioning on Action Units and TS~\cite{chan2018everybody} performs video retargeting conditioning on pose estimation.

\setlength\abovecaptionskip{-5pt}
\begin{figure*}[thb]
\includegraphics[width=1.0\textwidth]{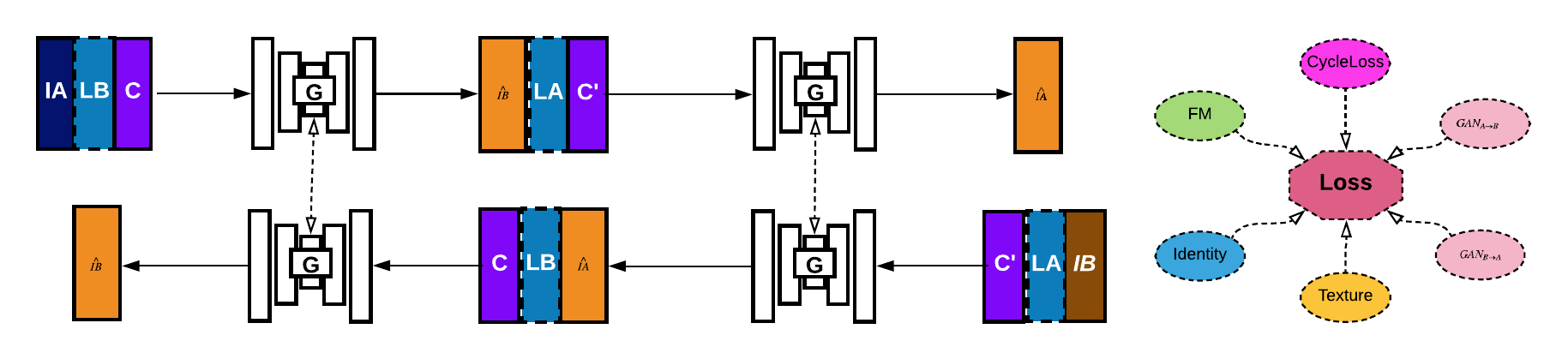}
\caption{Overview of training pipeline: In the forward cycle, original image $IA$ is first translated to $\widehat{IB}$ given target emotion $LB$ and modality $C$ and then mapped back to $\widehat{IA}$ given condition pair ($LA$,$C'$) encoding the original image. The backward cycle follows similar manner starting from $IB$ but with opposite condition encodings using the same generator $G$. Identity preservation and modality constraints are explicitly modeled in our loss design.}
\label{fig:overview}
\end{figure*}

\section{Proposed Method}
\label{sec:method}
\paragraph{Problem formulation} Given domains $\mathcal{X}_{1}, \mathcal{X}_{2}, \mathcal{X}_{3},... \mathcal{X}_{n}$ of different modalities, our goal is to learn a \textit{single} general mapping function
\begin{equation}
    \begin{split}
        G:\mathcal{X}_{i} \rightarrow \mathcal{X}_{j}, \forall  i,j\in \{1,2,3,... n\}
    \end{split}
    \label{eqn:G}
\end{equation}
that transforms $\mathcal{I}_{A}$ from domain $A$ to $\mathcal{I}_{B}$ from domain $B$ in a continuous manner (See Figure~\ref{fig:fail}). Eqn~\ref{eqn:G} implicitly implies that $G$ is bidirectional given desired conditions. We use facial landmark $\mathcal{L}_{j}\in R^{1\times H\times W}$ to denote facial expression in domain $j$. Facial expressions are represented as a vector of 2D keypoints with $N=68$, where each point $u_{i}=(x_{i}, y_{i})$ is the $i$th pixel location in $\mathcal{L}_{j}$. We use attribute vector $\overline{c}=[c_{1},c_{2},c_{3},...c_{n}]$ to represent the target domain. Formally, our input/output are tuples of the form $(\mathcal{I}_{A},\mathcal{L}_{B},c_{B})/ (\mathcal{I}_{B},\mathcal{L}_{A},c_{A}) \in R^{(3+1+n)\times H\times W}$. 

\paragraph{Model architecture}The overall pipeline of our approach is straightforward, shown in Figure~\ref{fig:overview} consisting of three main components: (1) A generator $G(\mathcal{I}|\mathcal{L},\overline{c})$, which renders an input face in domain $\overline{c_{1}}$ to the same person in another domain $\overline{c_{2}}$ given conditional facial landmarks. $G$ is bidirectional and reused in both forward as well as backward cycle. First mapping $\mathcal{I}_{A}\rightarrow \widehat{\mathcal{I}_{B}}\rightarrow \widehat{\mathcal{I}_{A}}$ and then mapping back $\mathcal{I}_{B}\rightarrow \widehat{\mathcal{I}_{A}}\rightarrow \widehat{\mathcal{I}_{B}}$ given conditional pair $(\mathcal{L}_{B}, \overline{c_{B}})/(\mathcal{L}_{A}, \overline{c_{A}})$. (2) A set of discriminators $D_{i}$ at different levels of resolution that distinguish generated samples from real ones. Instead of mapping $\mathcal{I}$ to a single scalar which signifies ``real'' or ``fake'' , we adopt PatchGAN~\cite{zhu2017unpaired} which uses a fully convnet that outputs a matrix where each element $M_{i,j}$ represents the probability of overlapping patch $ij$ to be real. If we trace back to the original image, each output has a $70\times 70$ receptive field. 
(3)Our loss function that takes into account identity preservation and texture consistency between different domains. In the following sections, we elaborate on each module individually and then combine them together to construct PortraitGAN.

\subsection{Base Model}
\label{sec:baseline}
To begin with, we consider manipulation of emotions in the same domain, i.e. $\mathcal{I}_A$ and $\mathcal{I}_B$ are of same texture and style, but with different face shapes denoted by facial landmarks $\mathcal{L}_A$ and $\mathcal{L}_B$. Under this scenario, it's sufficient to incorporate only forward cycle and conditional modality vector is not needed. The adversarial loss conditioned on facial landmarks follows Eqn~\ref{eqn:uni_gan}. 

\begin{equation}
\begin{split}
\mathcal{L}_{\text{GAN}}(G,D) = E_{{\mathcal{I}_B}\sim p(\mathcal{I}_B)}[\log (D(\mathcal{I}_B)] + E_{(\mathcal{I}_A,\mathcal{I}_B)\sim p(\mathcal{I}_A,\mathcal{I}_B)} \\
[\log (1-D(G(\mathcal{I}_A,\mathcal{L}_B)))]
\end{split}
\label{eqn:uni_gan}
\end{equation}

\noindent A face verification loss is desired to preserve identity between $\mathcal{I}_B$ and $\widehat{\mathcal{I}_B}$. However in our experiments, we find $\ell_1$ loss to be enough and it's better than $\ell_2$ loss as it alleviates blurry output and acts as an additional regularization~\cite{isola2017image}.

\begin{equation}
    \mathcal{L}_{1}^{\text{identity}}(G) = E_{(\mathcal{I}_A,\mathcal{I}_B)\sim p(\mathcal{I}_A,\mathcal{I}_B)} ||\mathcal{I}_B - G(\mathcal{I}_A,\mathcal{L}_B) ||_1
\label{eqn:uni_l1}
\end{equation}

\noindent The overall loss is a combination of adversarial loss and $\ell_1$ loss, weighted by $\lambda$. We will later extend this model to Eqn~\ref{eqn:bi_model}.

\begin{equation}
    G^{*} = \arg\min_{G} \max_{D} \quad \mathcal{L}_{\text{GAN}}(G,D) + \lambda \mathcal{L}_1^{\text{identity}}(G)
    \label{eqn:opt1}
\end{equation}


\subsection{Multi-level Adversarial Supervision}
\label{sec:multi}
Manipulation at a landmark level requires high-resolution synthesis, which is notoriously challenging for generative adversarial networks~\cite{goodfellow2016nips}. This is because training the whole system consists of optimizing two individual networks, where each update in either component could change the entire equilibrium. 

Here we use two major strategies for improving generation quality and training stability. First is to provide additional constraints on the training process. On the one hand, our conditional facial landmark acts as a constraint for generation. On the other hand, we adopt multi-level feature matching loss~\cite{gatys2016image,salimans2016improved} to explicitly require $G$ to match statistics of real data that $D$ finds most discriminative,

\begin{equation}
 \mathcal {L}_{FM}(G,D_{k})=\| \mathbb { E }_{\mathbf{\mathcal{I}_{B}}}  D_{k} ( \mathbf{\mathcal{I}_{B}} ) - \mathbb { E }_{{ \mathbf{\mathcal{I}_{A}}},\mathbf{\mathcal{I}_{B}}} 
 D_{k} ( G ( \mathbf{\mathcal{I}_{A}},\mathbf{\mathcal{L}_{B}} ) ) \| _ { { 1} } 
\label{eqn:feat_match}
\end{equation}

\noindent where $D$ acts like a feature extraction function that ``passes'' its strong feature representation to relatively weak generator $G$. $\mathcal{I}_{B}$ is a real face randomly chosen from pool that queues authentic samples for reducing sample variance, similar in spirit to experience replay buffer in DQN training~\cite{mnih2015human}. 

\vspace{-2mm}
\setlength\abovecaptionskip{-5pt}
\begin{figure}[thb]
\includegraphics[width=0.5\textwidth]{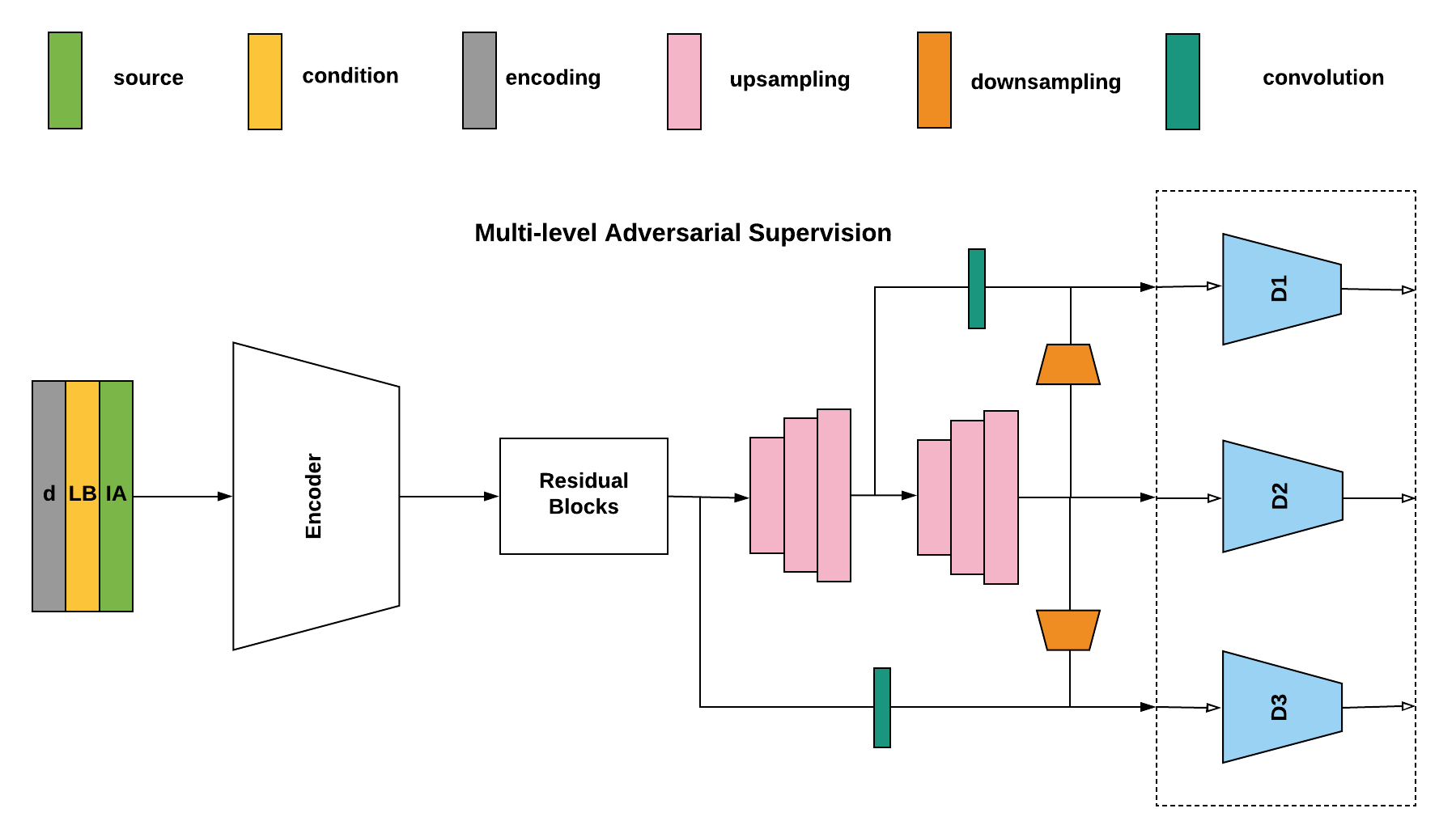}
\caption{Multi-level adversarial supervision}
\vspace{-3mm}
\label{fig:multi-level}
\end{figure}

Our second strategy is to provide fine-grained guidance by propagating multi-level features for adversarial supervision. Cascaded upsampling layers in $G$ are connected with auxiliary convolutional branches to provide images at different scales ($\widehat{\mathcal{I}_{B1}},\widehat{\mathcal{I}_{B2}},\widehat{\mathcal{I}_{B3}}...\widehat{\mathcal{I}_{Bm}}$), where $m$ is the number of upsampling blocks. Images generated at the intermediate stage $i$, together with corresponding downsampled images from the last stage, are fed into discriminator $D_{i}$, which is trained to classify real samples from generated ones through minimizing the following loss,

\begin{equation}
\begin{split}
    \mathcal { L } _ { D _ { i } } = - \frac { 1} { 2} \mathbb { E } _ { x _ { i }  } [ \log D _ { i } ( x _ { i } ) ] - \frac { 1} { 2} \mathbb { E } _ { g _ { i }  } [ \log ( 1 - D _ { i } ( g _ { i } ))]
\end{split}
    \label{eqn:multi-level}
\end{equation}

\begin{equation}
    G^{*}=\min _ { G } \max _ { D _ { k, k\epsilon I(k)}} \sum _ { k } \mathcal { L } _ 
    {\text{GAN} } \left( G ,D _ { k } \right) + \lambda \mathcal{L}_1^{\text{identity}}(G)
    \label{eqn:opt2}
\end{equation}

\noindent where $x_{i}$ is sampled from real distribution $p_{data_{i}}$ and $g_{i}$ from model distribution at scale $i$. $I(k)$ indicates all possible values of $k$. The auxiliary branches at different stages of generation provide more gradient signals for training the whole network, hence multi-level adversarial supervision. Compared to~\cite{Han17stackgan2},  our discriminators responsible for different levels are optimized as a whole rather than individually for each level. The increased discriminative ability from $D$ in turn provides further guidance when training $G$ and the two are alternatively optimized until convergence (Eqn~\ref{eqn:opt2}). 

\subsection{Texture consistency}
\label{sec:texture}
When translating between different modalities in high-resolution, texture differences become easy to observe. Inspired by~\cite{gatys2016image}, we let $\psi_{\mathcal{I},L}^{k}$ be the vectorized $k$th extracted feature map of image $\mathcal{I}$ from neural network $\psi$ at layer $L$. $\mathcal{G}_{\mathcal{I},L} \in R^{\kappa \times \kappa}$ is defined as,

\begin{equation}
    \mathcal{G}_{\mathcal{I},L}(k,l)=<\psi_{\mathcal{I},L}^{k}, \psi_{\mathcal{I},L}^{l}>=\sum_{i}\psi_{\mathcal{I},L}^{k}(i)\cdot \psi_{\mathcal{I},L}^{l}(i)
    \label{eqn:gram}
\end{equation}

\noindent where $\kappa$ is the number of feature maps at layer $L$ and $\psi_{\mathcal{I},L}^{k}(i)$ is $i$th element in the feature vector.
Eqn~\ref{eqn:gram} also known as Gram matrix can be seen as a measure of the correlation between feature maps $k$ and $l$, which only depends on the number of feature maps, not the size of $\mathcal{I}$. For image $\mathcal{I}_A$ and $\mathcal{I}_B$, the texture loss at layer $L$ is,

\begin{equation}
\begin{split}
    \mathcal{L}_{texture}^{L}(\mathcal{I}_A,\mathcal{I}_B)=||\mathcal{G}_{\mathcal{I}_A,L}-\mathcal{G}_{\mathcal{I}_B,L}||^{2}  \\ =\sum_{k,l}(\psi_{\mathcal{I}_A,L}(k,l)-\psi_{\mathcal{I}_B,L}(k,l))^{2}
\end{split}
\label{eqn:texture}
\end{equation}

\noindent We obtain obvious improvement in quality of texture in cross-modality manipulation during evaluation and we use pretrained VGG19 in our experiments with its parameters frozen during updates. 

\subsection{Going Beyond: Bidirectional Transfer}
\label{sec:bidirection}
Bringing all pieces together, we are now ready to extend our Base Model (Eqn~\ref{eqn:uni_gan}) to PortraitGAN  by incorporating bidirectional mapping and conditional vector $\overline{c}$, which represents the target domain. Eqn~\ref{eqn:uni_gan} now becomes,

\begin{equation}
    \begin{split}
    \mathcal{L}_{\text{GAN}_{A\rightarrow B}}(G,D) = E_{{\mathcal{I}_B}\sim p(\mathcal{I}_B)}[\log(D(\mathcal{I}_B))] + \\ E_{(\mathcal{I}_A,\mathcal{I}_B)\sim p(\mathcal{I}_A,\mathcal{I}_B), \overline{c}\epsilon I(\overline{c})} 
[\log(1-D(G(\mathcal{I}_A,\mathcal{L}_B,\overline{c})))]
    \end{split}
\label{eqn:bi_model}
\end{equation}

\noindent Forward cycle and backward cycle encourages one-to-one mapping from different modalities and thus helps preserve identity,

\begin{equation}
\begin{split}
    \mathcal{L}_{\text{cyc}}(G) = E_{\mathcal{I}_A,\mathcal{L}_B,\overline{c},\overline{c}'}[||G(G(\mathcal{I}_A,\mathcal{L}_B,\overline{c}),\mathcal{L}_A,\overline{c}')-\mathcal{I}_A||]_1  \\
    +E_{\mathcal{I}_B,\mathcal{L}_A,\overline{c},\overline{c}'}[||G(G(\mathcal{I}_B,\mathcal{L}_A,\overline{c}'),\mathcal{L}_B,\overline{c})-\mathcal{I}_B||]_1
\end{split}
\end{equation}

\noindent where $\overline{c}$ and $\overline{c}'$ encodes different modalities. Therefore, only one set of generator/discriminator is used for bidirectional manipulation. We find that both forward and backward cycle are essential for translation between domains, which is consistent with observation in~\cite{zhu2017unpaired}. $\mathcal{L}_{\text{GAN}_{B\rightarrow A}}$ can be written in a similar fashion and below is our full objective,

\begin{equation}
\begin{split}
    \mathcal{L}_{\text{PortraitGAN}} = \sum\mathcal{L}_{\text{GAN}_{A\rightarrow B}} + \sum\mathcal{L}_{\text{GAN}_{B\rightarrow A}}+  \alpha *\mathcal{L}_{\text{cyc}} \\
    + \beta * \sum\mathcal{L}_{\text{FM}} + \gamma * \mathcal{L}_{\text{identity}} + \eta * \mathcal{L}_{\text{texture}}
\end{split}
\label{eqn:obj}
\end{equation}

\noindent where $\alpha$, $\beta$, $\gamma$, $\eta$ controls the weight of cycle-consistency loss, feature matching loss, identity loss and texture loss respectively.


\section{Experimental Evaluation}
\label{sec:eval}
This section provides a thorough evaluation of our framework. We test our model's two main components: continuous editing and multi-modality transformation and compare our model with competing techniques in terms of flexibility and perceptual quality by performing both qualitative and quantitative studies. 
Note that considering the lack of groundtruth data for many face manipulation tasks, we leverage the result of~\cite{johnson2016perceptual} to generate pseudo-targets for multi-modality manipulation, but our framework can be extended to any desired domains.

\paragraph{Implementation Details} Each training step takes as input a tuple of four images $(\mathcal{I}_{A}$, $\mathcal{I}_{B}$, $\mathcal{L}_{A}$, $\mathcal{L}_{B})$ randomly chosen from possible modalities of the same identity. Attribute conditional vector, represented as a one-hot vector, is replicated spatially before channel-wise concatenation with corresponding image and facial landmarks. Our generator uses 4 stride-2 convolution layers, followed by 9 residual blocks and 4 stride-2 transpose convolutions while auxiliary branch uses one-channel convolution for fusion of channels. We use two 3-layer PatchGAN~\cite{zhu2017unpaired} discriminators for multi-level adversarial supervision and Least Square loss~\cite{mao2017least} for stable training. Layer conv$1\_1$-conv$5\_1$ of VGG19~\cite{simonyan2014very} are used for computing texture loss. We set $\alpha$, $\beta$, $\gamma$ , $\eta$ as 2, 10, 5, 10 for evaluation. The training time for PortraitGAN takes about 50 hours on a single Nvidia 1080 GPU. 


\vspace{-2mm}
\paragraph{Dataset} \textit{Training and validation:} The Radboud Faces Database~\cite{langner2010presentation} contains 4,824 images with 67 participants, each performing 8 canonical emotional expressions: anger, disgust, fear, happiness, sadness, surprise, contempt, and neutral. iCV Multi-Emotion Facial Expression Dataset~\cite{lusi2017joint} is designed for micro-emotion recognition (5184x3456 resolution), which includes 31,250 facial expressions performing 50 different emotions. \textit{Testing:} We collect 20 videos of high-resolution from Youtube (abbreviated as HRY Dataset) containing people giving speech or address for testing. For the above datasets, we use dlib~\cite{dlib09} for facial landmark extraction and~\cite{johnson2016perceptual} for generating portraits of multiple styles. Note that during testing, groundtruths are used only for evaluation purposes.

\subsection{Quantitative Evaluation}
\label{sec:quaneval}
In this section, we first compare our framework against two state-of-the-art in the task of modality transformation. 

\begin{table}[!ht]
\begin{center}
\begin{tabular}{l c c c}
Method & MSE$\downarrow$ &  SSIM$\uparrow$ &  inference time(s)$\downarrow$ \\
\hline
CycleGAN &  0.028 &   0.473  &  0.365 \\
StarGAN  &  0.029  &  0.483  &  \textbf{0.277}\\
\hline
\textbf{Ours}    &  \textbf{0.025}  &  \textbf{0.517}  &  0.290 \\
\hline
\end{tabular}
\end{center}
\vspace{2mm}
\caption{Quantitative evaluation on modality manipulation task. For MSE, the lower the better, SSIM the higher the better. Ours achieve the best score while maintaining fast inference speed.}
\label{tab:metric}
\end{table}

\vspace{-2mm}
\paragraph{Evaluation metrics}
CycleGAN~\cite{zhu2017unpaired} first proposes cycle-consistency to improve sample-efficiency of pix2pix~\cite{isola2017image} and is one of the best models in two-domain translation. StarGAN~\cite{choi2017stargan} adopts a classifier for supervision and is capable of translation across domains. For fair comparison, we retrain 512x512 version of CycleGAN and StarGAN with domain dimension set as two and randomly choose 368 images from HRY dataset with different identities and expressions for natural to single stylized modality evaluation. We fix extracted landmarks unchanged during evaluation for PortraitGAN. 

\begin{table}[!ht]
\begin{center}
\begin{tabular}{l c c c}
Method (\%) & 1st round   &  2nd round  & Average \\
\hline
StarGAN & 31.2 & 32.3  &  31.75  \\
CycleGAN &32.0 & 32.5 &  32.25  \\
\hline
\textbf{Ours}   &  \textbf{36.8} & \textbf{35.2}  &  \textbf{36.0}  \\
\hline 
\end{tabular}
\end{center}
\vspace{2mm}
\caption{Subjective ranking for different models based on perceptual evaluation of modality manipulation performance.}
\label{tab:study}
\end{table}

\vspace{-2mm}
\paragraph{Subjective user study}
We also conduct human subjective study as in~\cite{isola2017image,zhu2017unpaired,wang2017high,choi2017stargan} on performance of natural to single stylized modality manipulation. We collect responses from 10 users (5 experts, 5 non-experts) based on their preferences about images displayed at each group in terms of perceptual realism and identity preservation. Each group consists of one photo input and three randomly shuffled manipulated images generated by cycleGAN~\cite{zhu2017unpaired}, StarGAN~\cite{choi2017stargan} and our proposed approach with landmarks unchanged. We conducted two rounds of user study where the 1st round has a time limit of 5 seconds while 2nd round is unlimited. There are in total 100 images and each user is asked to rank three methods on each image twice. Our model gets the best score among three methods as shown in Table~\ref{tab:study}.

\subsection{Qualitative Evaluation}
\label{sec:quaeval}
In this section, we conduct ablation study and validate the effectiveness of our design in continuous editing. We also compare against state-of-the-art generative models on tasks of continuous shape editing and simultaneous shape and modality manipulations. Finally, we show some manipulation cases using our developed interactive interface.

\begin{figure}[htb]
  \hbox{ \hspace{-4mm} \includegraphics[width=0.5\textwidth]{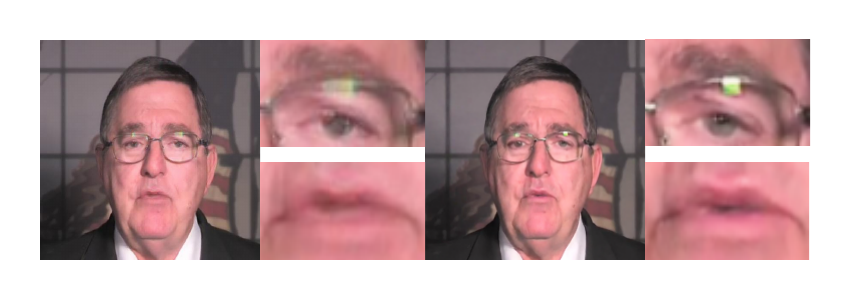}}
    \caption{Effect of multi-level adversarial supervision. Left/Right: wo/w multi-level adversarial supervision. }
    \label{fig:ablation}
\end{figure}
\vspace{-2mm}

\begin{figure}[htb]
    \hbox{\hspace{-4mm}
    \includegraphics[width=0.5\textwidth]{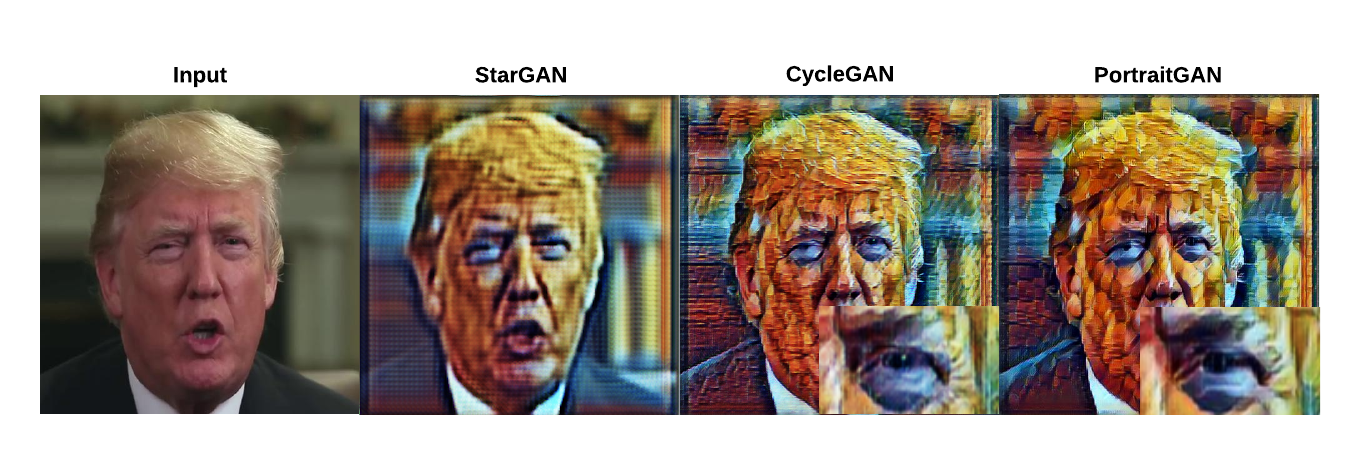}}
    \caption{Perceptual quality comparison against StarGAN and CycleGAN in terms of modality transformation. }
    \label{fig:comp}
\end{figure}

\paragraph{Ablation study} Each component is crucial for the proper performance of the system. $\mathcal{L}_{cyc}$ and $\mathcal{L}_{identity}$ for identity preservation between modalities, $\mathcal{L}_{FM}$ and $\mathcal{L}_{GAN}$ for high-resolution generation, $\mathcal{L}_{texture}$ for modality transformation. Removing any of these elements would damage our network. For example, 
Figure~\ref{fig:ablation} shows the effect of multi-level adversarial supervision. As can be seen, generated result with our component displays better perceptual quality with more high-frequency details. Texture quality would be compromised without texture loss (Figure~\ref{fig:comp}). Last but not least, bidirectional cycle-consistency eliminates the need of classifier used in~\cite{choi2017stargan} for multi-domain manipulation. 

\begin{figure}[htb]
  \hbox{ \hspace{-4mm} \includegraphics[width=0.5\textwidth]{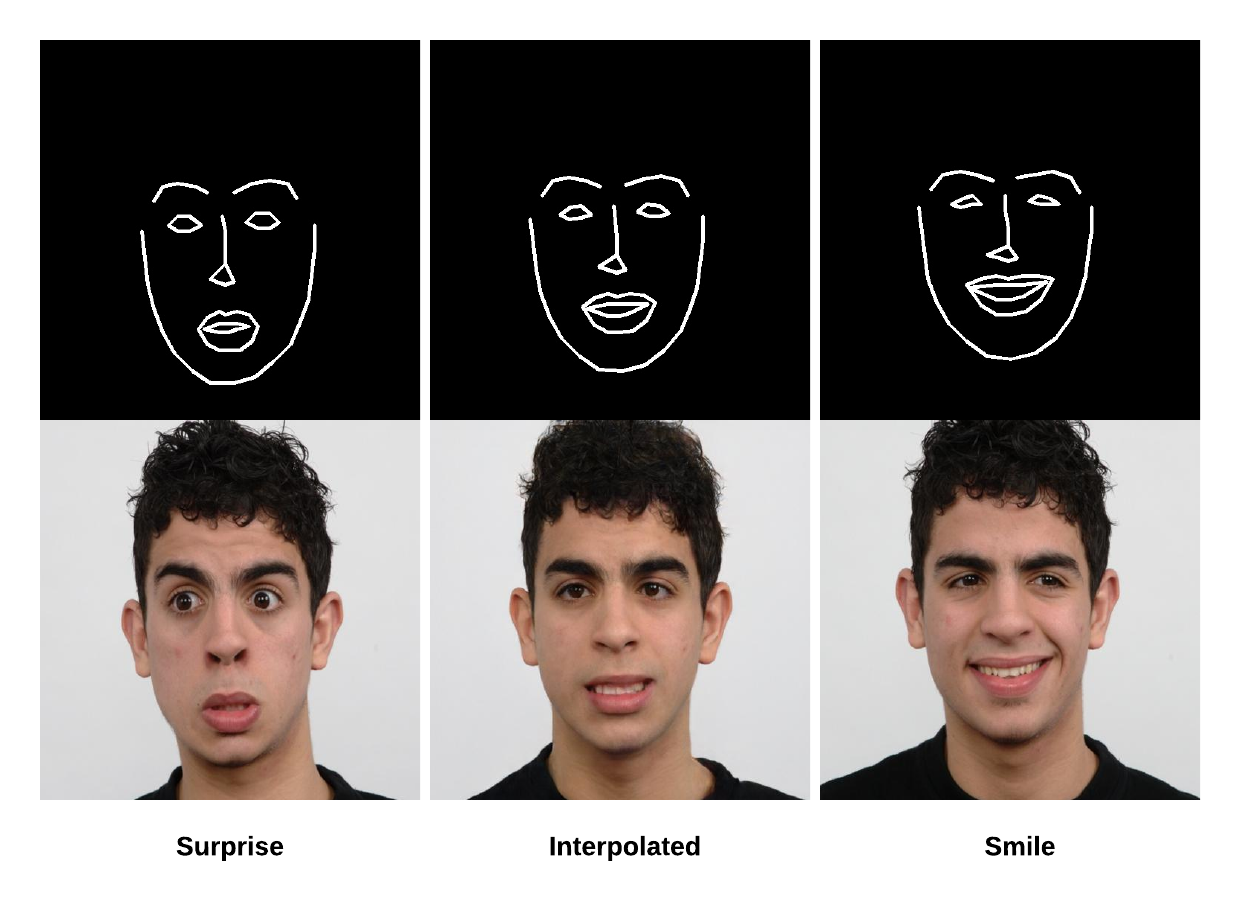}}
    \caption{Our model is able to interpolate expressions that are beyond canonical discrete expressions in the training set. }
    \label{fig:interp}
\end{figure}
\vspace{-2mm}

\begin{figure}[!htb]
    \hspace{-5mm}
    \includegraphics[width=0.52\textwidth]{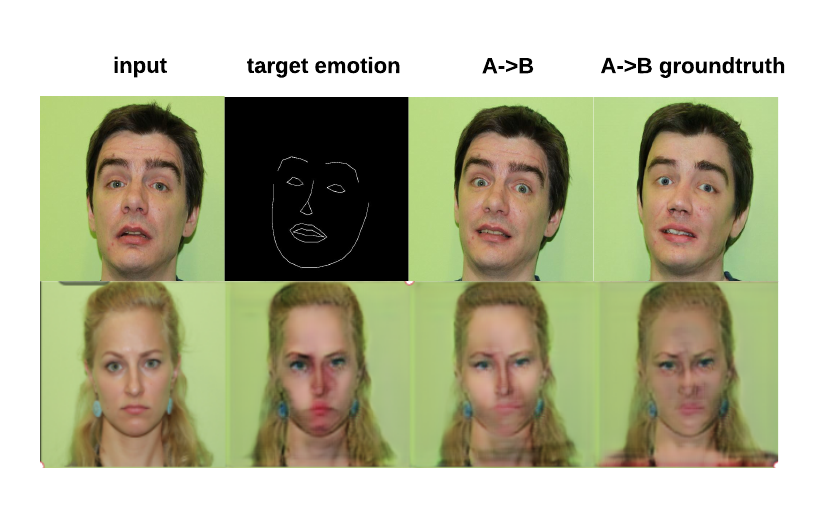}
    \caption{Comparison with StarGAN in the task of continuous shape editing. Our model outperforms StarGAN for manipulations among subtle motion variations on iCV dataset (2nd row vs 3rd row). }
    \label{fig:p2p}
\end{figure}

\begin{figure*}[!htb]
    \centering
    \includegraphics[width=1.0\textwidth]{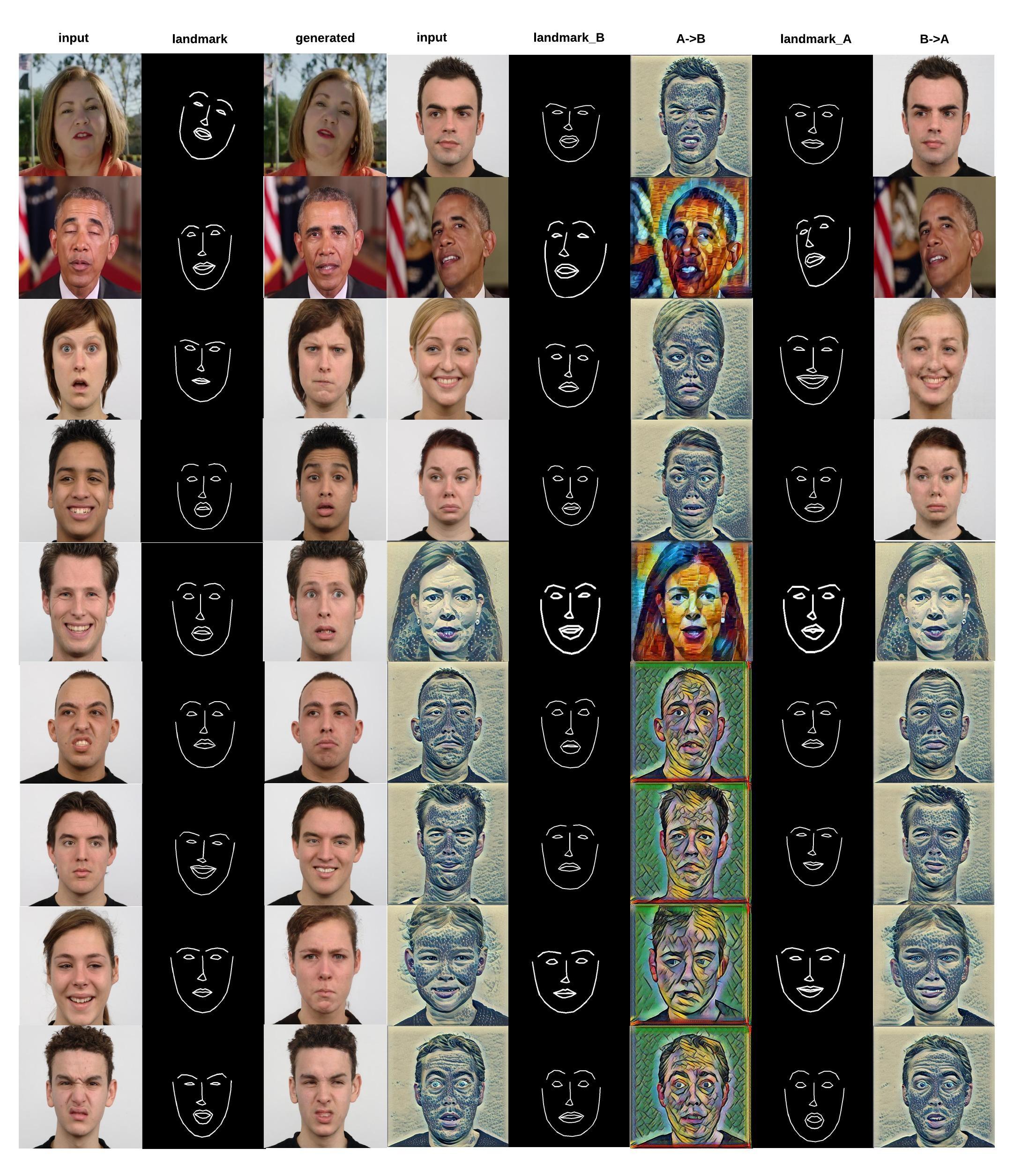}
    \caption{More results for continuous shape edits and simultaneous shape and modality manipulation results by PortraitGAN. }
    \label{fig:p2s}
\end{figure*}

\vspace{-2mm}
\paragraph{Continuous shape editing}
Figure~\ref{fig:interp} shows interpolated expression of our model on Rafd, which is beyond its original 8 canonical expressions. Note that CycleGAN can't transfer in the same domain. On iCV dataset, we train StarGAN on 50 discrete micro emotions, but it collapsed. Perhaps it's because StarGAN requires strong classification loss for supervision, which is hard to obtain on iCV dataset. On the other hand, our model successfully operates in the continuous space that captures subtle variations of face shapes (Figure~\ref{fig:p2p}). Another intriguing fact we observed is that boundary width of landmark doesn't have obvious influence on output. More results are available in Figure~\ref{fig:p2s} (column 1-3).

\paragraph{Simultaneous shape and modality manipulation}
\label{sec:p2s} Simultaneous shape and modality manipulations on HRY dataset is shown in 
Figure~\ref{fig:p2s} (column 4-8). If look closely, our model is capable of hallucinating teeth (1st row) and capturing details such as ear rings (5th row), If landmark is fixed, our model then acts like a modality transfer model except that it can achieve bidirectional modality transfer with a minor change of attribute conditional vector $\overline{c}$. 

To compare our approach with CycleGAN~\cite{zhu2017unpaired} and StarGAN~\cite{choi2017stargan}, we use the following pipeline: Given image pair \{$\mathcal{I}_A$,$\mathcal{I}_B$\}, which are from domain $A$ and $B$, CycleGAN translates $\mathcal{I}_A$ to $\mathcal{T}_A$, which has content from $\mathcal{I}_A$ and modality from $\mathcal{I}_B$. This can be achieved with our approach but with landmark $\mathcal{L}_A$ unchanged. Similarly, we treat modalities as visual attributes and train StarGAN accordingly. Figure~\ref{fig:comp} shows the result of four models. As can be seen, ours is much sharper and perceptually coherent.

\begin{figure}[!bht]
    \hspace{-3mm}
    \includegraphics[width=0.5\textwidth]{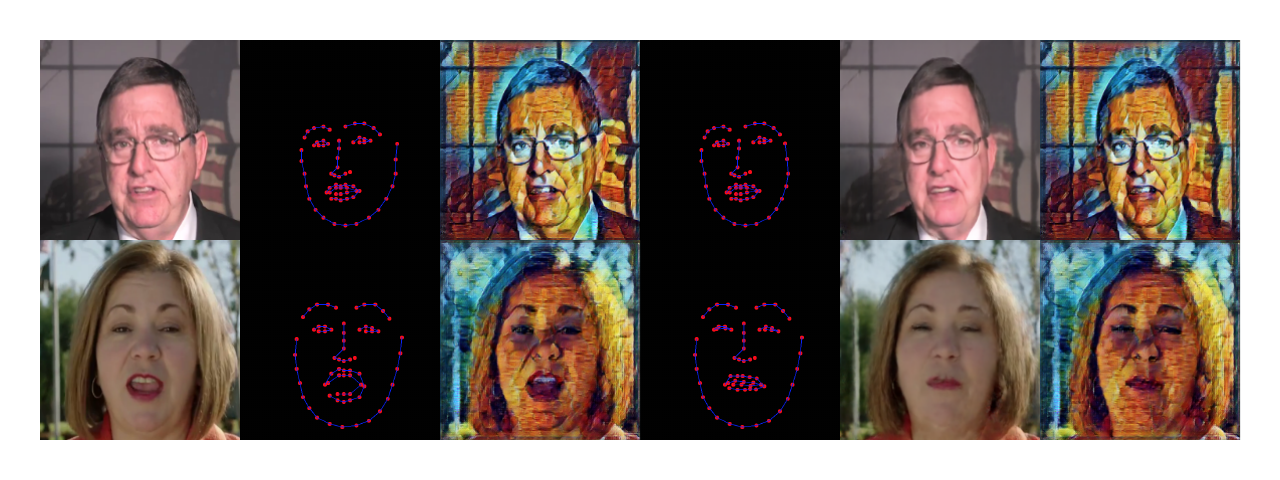}
    \caption{Interactive manipulation without constraints. Column 1st-3rd: Modality conversion given input and auto-detected landmarks;  3rd-5th: Simultaneous inverse-modality and attribute conversion; 4th-6th: Modality conversion. 1st row conducts face-slimming while 2nd row closes both eyes and mouth, demonstrating that our model is able to generalize to emotions not present in training set.}
    \label{fig:fail}
\end{figure}

\begin{figure}[!htb]
    \hspace{-4mm}
    \includegraphics[width=0.51\textwidth]{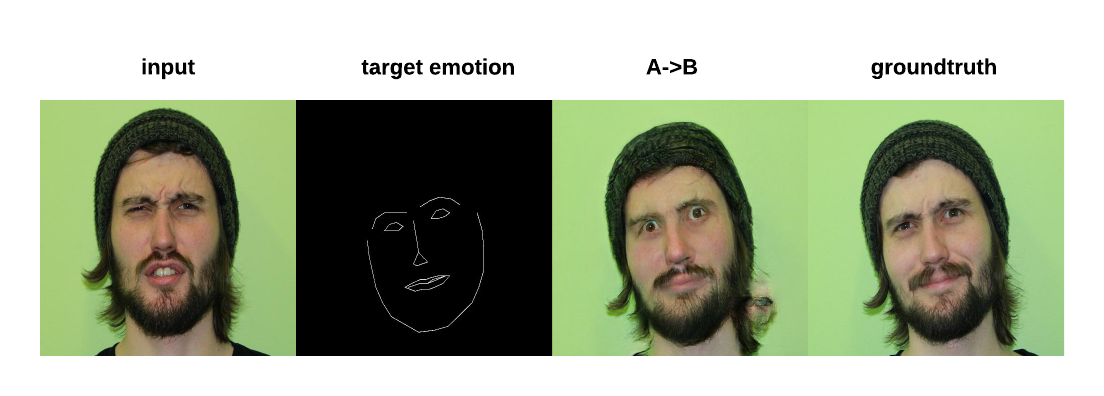}
    \caption{Failure cases: The reason could be that facial landmarks don't capture well enough details of micro-emotions.}
    \label{fig:fail2}
\end{figure}

\vspace{-3mm}
\paragraph{Interactive manipulation in the wild} Compared to discrete conditional labels, facial landmark gives full freedom for continuous shape editing. 
To test the limit of our model, we develop an online interactive editing interface, where users can manipulate facial landmarks manually and evaluate the model directly. This proves to be more challenging than landmark interpolation, as these edits go far beyond normal expressions in the training set. Figure~\ref{fig:fail} shows some interesting results. As can be seen, our model can successfully perform simultaneous face-slimming and modality manipulation from input of the original modality. Figure~\ref{fig:fail2} shows a failure case when generated result doesn't conform to groundtruth.

\vspace{-2mm}
\setlength\abovecaptionskip{-5pt}
\vspace{-4mm}

\section{Conclusion}
\label{sec:discuss}
To our best knowledge, this is the first work that explores the possibility of combining emotion manipulation and modality transfer in a single framework. It advances current research, which so far, had only considered discrete manipulation in a single domain. We exhaustively validate our approach both qualitatively and quantitatively on iCV~\cite{langner2010presentation}, RaFD~\cite{lusi2017joint} as well as HRY datasets and conducted ablation study for each of the components proposed in our model. The results show the promise of extending our framework to embrace transformations beyond emotion and style, however given those data are limited, we leave it as future work. 

\clearpage
{\small
\bibliographystyle{aaai}
\bibliography{egbib}

\begin{thebibliography}{}

\bibitem[\protect\citeauthoryear{Averbuch-Elor \bgroup et al\mbox.\egroup
  }{2017}]{averbuch2017bringing}
Averbuch-Elor, H.; Cohen-Or, D.; Kopf, J.; and Cohen, M.~F.
\newblock 2017.
\newblock Bringing portraits to life.
\newblock {\em ACM Transactions on Graphics (TOG)} 36(6):196.

\bibitem[\protect\citeauthoryear{Blanz and Vetter}{1999}]{blanz1999morphable}
Blanz, V., and Vetter, T.
\newblock 1999.
\newblock A morphable model for the synthesis of 3d faces.
\newblock In {\em Proceedings of the 26th annual conference on Computer
  graphics and interactive techniques},  187--194.
\newblock ACM Press/Addison-Wesley Publishing Co.

\bibitem[\protect\citeauthoryear{Blanz \bgroup et al\mbox.\egroup
  }{2003}]{blanz2003reanimating}
Blanz, V.; Basso, C.; Poggio, T.; and Vetter, T.
\newblock 2003.
\newblock Reanimating faces in images and video.
\newblock In {\em Computer graphics forum}, volume~22,  641--650.
\newblock Wiley Online Library.

\bibitem[\protect\citeauthoryear{Chan \bgroup et al\mbox.\egroup
  }{2018}]{chan2018everybody}
Chan, C.; Ginosar, S.; Zhou, T.; and Efros, A.~A.
\newblock 2018.
\newblock Everybody dance now.
\newblock {\em arXiv preprint arXiv:1808.07371}.

\bibitem[\protect\citeauthoryear{Chen and Koltun}{2017}]{chen2017photographic}
Chen, Q., and Koltun, V.
\newblock 2017.
\newblock Photographic image synthesis with cascaded refinement networks.
\newblock In {\em The IEEE International Conference on Computer Vision (ICCV)},
  volume~1.

\bibitem[\protect\citeauthoryear{Choi \bgroup et al\mbox.\egroup
  }{2017}]{choi2017stargan}
Choi, Y.; Choi, M.; Kim, M.; Ha, J.-W.; Kim, S.; and Choo, J.
\newblock 2017.
\newblock Stargan: Unified generative adversarial networks for multi-domain
  image-to-image translation.
\newblock {\em arXiv preprint arXiv:1711.09020}.

\bibitem[\protect\citeauthoryear{Gatys, Ecker, and
  Bethge}{2016}]{gatys2016image}
Gatys, L.~A.; Ecker, A.~S.; and Bethge, M.
\newblock 2016.
\newblock Image style transfer using convolutional neural networks.
\newblock In {\em Computer Vision and Pattern Recognition (CVPR), 2016 IEEE
  Conference on},  2414--2423.
\newblock IEEE.

\bibitem[\protect\citeauthoryear{Goodfellow}{2016}]{goodfellow2016nips}
Goodfellow, I.
\newblock 2016.
\newblock Nips 2016 tutorial: Generative adversarial networks.
\newblock {\em arXiv preprint arXiv:1701.00160}.

\bibitem[\protect\citeauthoryear{Hoffman \bgroup et al\mbox.\egroup
  }{2017}]{hoffman2017cycada}
Hoffman, J.; Tzeng, E.; Park, T.; Zhu, J.-Y.; Isola, P.; Saenko, K.; Efros,
  A.~A.; and Darrell, T.
\newblock 2017.
\newblock Cycada: Cycle-consistent adversarial domain adaptation.
\newblock {\em arXiv preprint arXiv:1711.03213}.

\bibitem[\protect\citeauthoryear{Isola \bgroup et al\mbox.\egroup
  }{2017}]{isola2017image}
Isola, P.; Zhu, J.-Y.; Zhou, T.; and Efros, A.~A.
\newblock 2017.
\newblock Image-to-image translation with conditional adversarial networks.
\newblock {\em arXiv preprint}.

\bibitem[\protect\citeauthoryear{Johnson, Alahi, and
  Fei-Fei}{2016}]{johnson2016perceptual}
Johnson, J.; Alahi, A.; and Fei-Fei, L.
\newblock 2016.
\newblock Perceptual losses for real-time style transfer and super-resolution.
\newblock In {\em European Conference on Computer Vision},  694--711.
\newblock Springer.

\bibitem[\protect\citeauthoryear{Kemelmacher-Shlizerman, Suwajanakorn, and
  Seitz}{2014}]{kemelmacher2014illumination}
Kemelmacher-Shlizerman, I.; Suwajanakorn, S.; and Seitz, S.~M.
\newblock 2014.
\newblock Illumination-aware age progression.
\newblock In {\em Proceedings of the IEEE Conference on Computer Vision and
  Pattern Recognition},  3334--3341.

\bibitem[\protect\citeauthoryear{King}{2009}]{dlib09}
King, D.~E.
\newblock 2009.
\newblock Dlib-ml: A machine learning toolkit.
\newblock {\em Journal of Machine Learning Research} 10:1755--1758.

\bibitem[\protect\citeauthoryear{Langner \bgroup et al\mbox.\egroup
  }{2010}]{langner2010presentation}
Langner, O.; Dotsch, R.; Bijlstra, G.; Wigboldus, D.~H.; Hawk, S.~T.; and
  Van~Knippenberg, A.
\newblock 2010.
\newblock Presentation and validation of the radboud faces database.
\newblock {\em Cognition and emotion} 24(8):1377--1388.

\bibitem[\protect\citeauthoryear{Lassner, Pons-Moll, and
  Gehler}{2017}]{lassner2017generative}
Lassner, C.; Pons-Moll, G.; and Gehler, P.~V.
\newblock 2017.
\newblock A generative model of people in clothing.
\newblock {\em arXiv preprint arXiv:1705.04098}.

\bibitem[\protect\citeauthoryear{Lau \bgroup et al\mbox.\egroup
  }{2009}]{lau2009face}
Lau, M.; Chai, J.; Xu, Y.-Q.; and Shum, H.-Y.
\newblock 2009.
\newblock Face poser: Interactive modeling of 3d facial expressions using
  facial priors.
\newblock {\em ACM Transactions on Graphics (TOG)} 29(1):3.

\bibitem[\protect\citeauthoryear{Liao \bgroup et al\mbox.\egroup
  }{2017}]{liao2017visual}
Liao, J.; Yao, Y.; Yuan, L.; Hua, G.; and Kang, S.~B.
\newblock 2017.
\newblock Visual attribute transfer through deep image analogy.
\newblock {\em arXiv preprint arXiv:1705.01088}.

\bibitem[\protect\citeauthoryear{Liu, Breuel, and
  Kautz}{2017}]{liu2017unsupervised}
Liu, M.-Y.; Breuel, T.; and Kautz, J.
\newblock 2017.
\newblock Unsupervised image-to-image translation networks.
\newblock In {\em Advances in Neural Information Processing Systems},
  700--708.

\bibitem[\protect\citeauthoryear{L{\"u}si \bgroup et al\mbox.\egroup
  }{2017}]{lusi2017joint}
L{\"u}si, I.; Junior, J. C.~J.; Gorbova, J.; Bar{\'o}, X.; Escalera, S.;
  Demirel, H.; Allik, J.; Ozcinar, C.; and Anbarjafari, G.
\newblock 2017.
\newblock Joint challenge on dominant and complementary emotion recognition
  using micro emotion features and head-pose estimation: Databases.
\newblock In {\em Automatic Face \& Gesture Recognition (FG 2017), 2017 12th
  IEEE International Conference on},  809--813.
\newblock IEEE.

\bibitem[\protect\citeauthoryear{Ma \bgroup et al\mbox.\egroup
  }{2017}]{ma2017pose}
Ma, L.; Jia, X.; Sun, Q.; Schiele, B.; Tuytelaars, T.; and Van~Gool, L.
\newblock 2017.
\newblock Pose guided person image generation.
\newblock In {\em Advances in Neural Information Processing Systems},
  405--415.

\bibitem[\protect\citeauthoryear{Mao \bgroup et al\mbox.\egroup
  }{2017}]{mao2017least}
Mao, X.; Li, Q.; Xie, H.; Lau, R.~Y.; Wang, Z.; and Smolley, S.~P.
\newblock 2017.
\newblock Least squares generative adversarial networks.
\newblock In {\em 2017 IEEE International Conference on Computer Vision
  (ICCV)},  2813--2821.
\newblock IEEE.

\bibitem[\protect\citeauthoryear{Mnih \bgroup et al\mbox.\egroup
  }{2015}]{mnih2015human}
Mnih, V.; Kavukcuoglu, K.; Silver, D.; Rusu, A.~A.; Veness, J.; Bellemare,
  M.~G.; Graves, A.; Riedmiller, M.; Fidjeland, A.~K.; Ostrovski, G.; et~al.
\newblock 2015.
\newblock Human-level control through deep reinforcement learning.
\newblock {\em Nature} 518(7540):529.

\bibitem[\protect\citeauthoryear{Pumarola \bgroup et al\mbox.\egroup
  }{2018a}]{pumarola2018ganimation}
Pumarola, A.; Agudo, A.; Martinez, A.~M.; Sanfeliu, A.; and Moreno-Noguer, F.
\newblock 2018a.
\newblock Ganimation: Anatomically-aware facial animation from a single image.
\newblock {\em arXiv preprint arXiv:1807.09251}.

\bibitem[\protect\citeauthoryear{Pumarola \bgroup et al\mbox.\egroup
  }{2018b}]{pumarola2018unsupervised}
Pumarola, A.; Agudo, A.; Sanfeliu, A.; and Moreno-Noguer, F.
\newblock 2018b.
\newblock Unsupervised person image synthesis in arbitrary poses.
\newblock In {\em Proceedings of the IEEE Conference on Computer Vision and
  Pattern Recognition},  8620--8628.

\bibitem[\protect\citeauthoryear{Reed \bgroup et al\mbox.\egroup
  }{2016}]{reed2016learning}
Reed, S.~E.; Akata, Z.; Mohan, S.; Tenka, S.; Schiele, B.; and Lee, H.
\newblock 2016.
\newblock Learning what and where to draw.
\newblock In {\em Advances in Neural Information Processing Systems},
  217--225.

\bibitem[\protect\citeauthoryear{Salimans \bgroup et al\mbox.\egroup
  }{2016}]{salimans2016improved}
Salimans, T.; Goodfellow, I.; Zaremba, W.; Cheung, V.; Radford, A.; and Chen,
  X.
\newblock 2016.
\newblock Improved techniques for training gans.
\newblock In {\em Advances in Neural Information Processing Systems},
  2234--2242.

\bibitem[\protect\citeauthoryear{Siarohin \bgroup et al\mbox.\egroup
  }{2017}]{siarohin2017deformable}
Siarohin, A.; Sangineto, E.; Lathuiliere, S.; and Sebe, N.
\newblock 2017.
\newblock Deformable gans for pose-based human image generation.
\newblock {\em arXiv preprint arXiv:1801.00055}.

\bibitem[\protect\citeauthoryear{Simonyan and
  Zisserman}{2014}]{simonyan2014very}
Simonyan, K., and Zisserman, A.
\newblock 2014.
\newblock Very deep convolutional networks for large-scale image recognition.
\newblock {\em arXiv preprint arXiv:1409.1556}.

\bibitem[\protect\citeauthoryear{Sucontphunt \bgroup et al\mbox.\egroup
  }{2008}]{sucontphunt2008interactive}
Sucontphunt, T.; Mo, Z.; Neumann, U.; and Deng, Z.
\newblock 2008.
\newblock Interactive 3d facial expression posing through 2d portrait
  manipulation.
\newblock In {\em Proceedings of graphics interface 2008},  177--184.
\newblock Canadian Information Processing Society.

\bibitem[\protect\citeauthoryear{Thies \bgroup et al\mbox.\egroup
  }{2016}]{thies2016face2face}
Thies, J.; Zollhofer, M.; Stamminger, M.; Theobalt, C.; and Nie{\ss}ner, M.
\newblock 2016.
\newblock Face2face: Real-time face capture and reenactment of rgb videos.
\newblock In {\em Proceedings of the IEEE Conference on Computer Vision and
  Pattern Recognition},  2387--2395.

\bibitem[\protect\citeauthoryear{Walker \bgroup et al\mbox.\egroup
  }{2017}]{walker2017pose}
Walker, J.; Marino, K.; Gupta, A.; and Hebert, M.
\newblock 2017.
\newblock The pose knows: Video forecasting by generating pose futures.
\newblock In {\em 2017 IEEE International Conference on Computer Vision
  (ICCV)},  3352--3361.
\newblock IEEE.

\bibitem[\protect\citeauthoryear{Wang \bgroup et al\mbox.\egroup
  }{2017}]{wang2017high}
Wang, T.-C.; Liu, M.-Y.; Zhu, J.-Y.; Tao, A.; Kautz, J.; and Catanzaro, B.
\newblock 2017.
\newblock High-resolution image synthesis and semantic manipulation with
  conditional gans.
\newblock {\em arXiv preprint arXiv:1711.11585}.

\bibitem[\protect\citeauthoryear{Yi \bgroup et al\mbox.\egroup
  }{2017}]{yi2017dualgan}
Yi, Z.; Zhang, H.; Tan, P.; and Gong, M.
\newblock 2017.
\newblock Dualgan: Unsupervised dual learning for image-to-image translation.
\newblock {\em arXiv preprint}.

\bibitem[\protect\citeauthoryear{Zhang \bgroup et al\mbox.\egroup
  }{2017}]{Han17stackgan2}
Zhang, H.; Xu, T.; Li, H.; Zhang, S.; Wang, X.; Huang, X.; and Metaxas, D.
\newblock 2017.
\newblock Stackgan++: Realistic image synthesis with stacked generative
  adversarial networks.
\newblock {\em arXiv: 1710.10916}.

\bibitem[\protect\citeauthoryear{Zhu \bgroup et al\mbox.\egroup
  }{2017a}]{zhu2017unpaired}
Zhu, J.-Y.; Park, T.; Isola, P.; and Efros, A.~A.
\newblock 2017a.
\newblock Unpaired image-to-image translation using cycle-consistent
  adversarial networks.
\newblock {\em arXiv preprint arXiv:1703.10593}.

\bibitem[\protect\citeauthoryear{Zhu \bgroup et al\mbox.\egroup
  }{2017b}]{zhu2017toward}
Zhu, J.-Y.; Zhang, R.; Pathak, D.; Darrell, T.; Efros, A.~A.; Wang, O.; and
  Shechtman, E.
\newblock 2017b.
\newblock Toward multimodal image-to-image translation.
\newblock In {\em Advances in Neural Information Processing Systems},
  465--476.

\end{thebibliography}
}

\end{document}